\documentclass[]{article}

\usepackage[]{fontenc}
\usepackage[utf8]{inputenc}
\usepackage[english]{babel}

\usepackage{natbib}
\usepackage[hidelinks]{hyperref}

\usepackage{booktabs}
\usepackage{tabularx}
\usepackage{siunitx}
\usepackage{multirow}
\usepackage{caption}

\usepackage{amsmath,amssymb}

\usepackage{graphicx}
\usepackage{placeins}

\title{Exploring the Dimensions of a Variational Neuron}

\author{
  Yves Ruffenach \\
  Conservatoire National des Arts et M\'etiers \\
  \texttt{yves@ruffenach.net} \\
  \href{https://orcid.org/0009-0009-4737-0555}{ORCID: 0009-0009-4737-0555}
}

\date{}

\begin{document}

\maketitle

\begin{abstract}
We introduce \emph{EVE} (\emph{Elemental Variational Expanse}), a variational distributional neuron formulated as a local probabilistic computational unit with an explicit prior, an amortized posterior and unit-level variational regularization. In most modern architectures, uncertainty is handled through global latent variables or parameter uncertainty, while the computational unit itself remains scalar. EVE instead relocates probabilistic structure to the neuron level and makes it locally observable and controllable.

The term \emph{dimensions} refers first and foremost to the neuron's internal latent dimensionality, denoted by \(k\). We study how varying \(k\), from the atomic case \(k=1\) to higher-dimensional latent spaces, changes the neuron's learned operating regime. We then examine how this primary axis interacts with two additional structural properties: local capacity control and temporal persistence through a neuron-level autoregressive extension.

To support this study, EVE is instrumented with internal diagnostics and constraints, including effective KL, a target band on \(\mu^2\), out-of-band fractions and indicators of drift and collapse. Across selected forecasting and tabular settings, we show that latent dimensionality, control, and temporal extension shape the neuron's internal regime, and that some neuron-level variables are measurable, informative, and related to downstream behavior. The paper provides an experimentally grounded first map of the design space opened by a variational neuron.
\end{abstract}

\section{Introduction}

Modern architectures, especially in sequential generation, organize causality and computation largely in symbolic space: attention, implicit recurrence and token-by-token decoding. In such systems, latent variables---when present---often remain auxiliary, while the effective dynamics is carried by a mostly deterministic decoder. More broadly, probabilistic learning provides principled tools for representing variation and uncertainty, but uncertainty is typically placed either in global latent states or in model parameters, as in Bayesian weights.

A structural asymmetry remains: most networks propagate scalars at the level of units, while uncertainty is handled above the unit or inside the weights. This limits what can be observed and controlled at the level of the computational building block itself. One may calibrate a global output, but one rarely observes---and even more rarely regulates---a local probabilistic activity. This motivates the idea of a neuron whose internal state is a distribution rather than a scalar.

Our goal is to study a variational neuron as an object with explicit internal degrees of freedom. The term \emph{dimensions} refers first and foremost to the possible values of the neuron's internal latent dimensionality, denoted by \(k\). We explore the same variational neuron from the atomic case \(k=1\) to higher-dimensional latent spaces and ask how expressivity, stability and internal behavior evolve with \(k\). We then examine how \(k\) interacts with two additional structural properties: local capacity control and temporal persistence.

From this perspective, we introduce \emph{EVE} (\emph{Elemental Variational Expanse}), a variational distributional neuron: a local probabilistic building block with an internal latent of dimension \(k\), an explicit prior, an amortized posterior and unit-level variational regularization. The unit is made observable and controllable through local constraints and diagnostics, including effective KL, a target band on \(\mu^2\), out-of-band fractions and indicators of drift and collapse.

\begin{figure}[htbp]
  \centering
  \includegraphics[width=\linewidth]{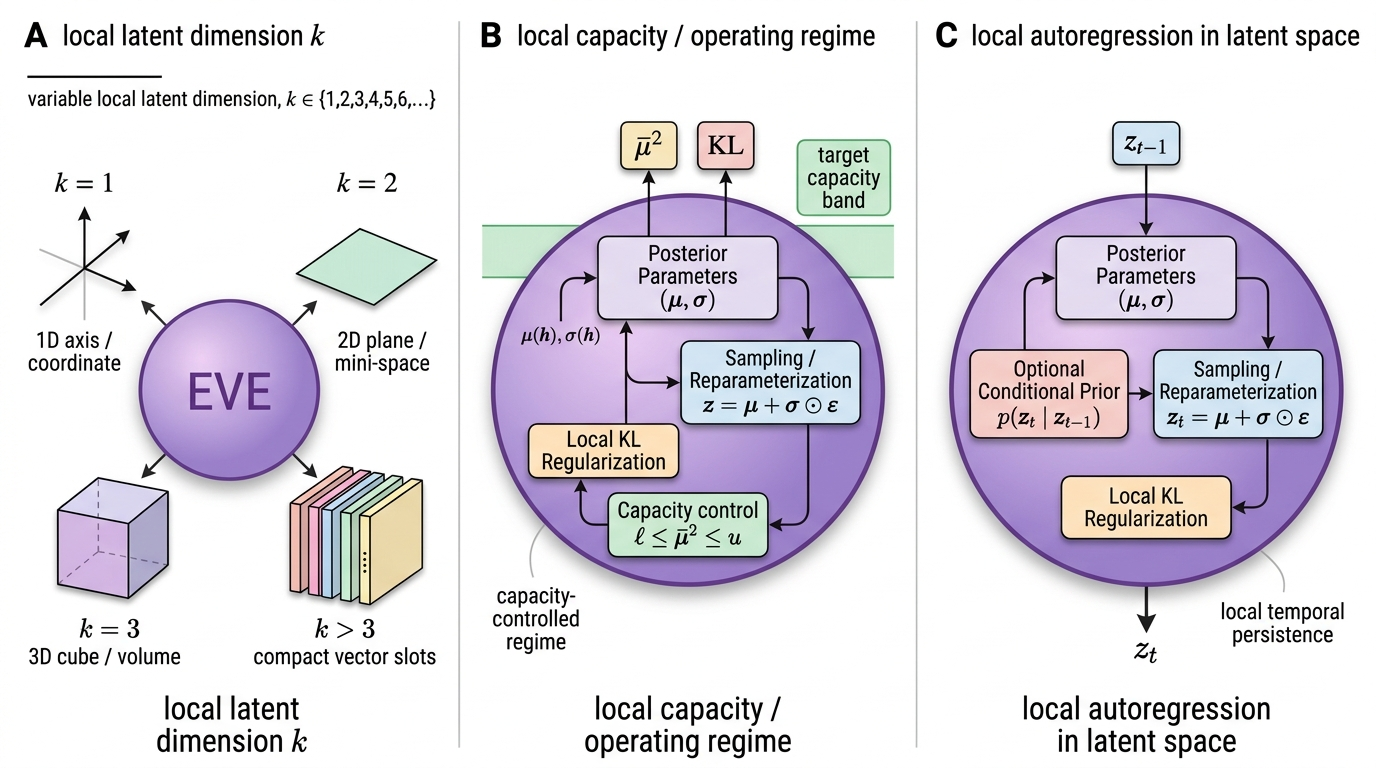}
\caption{\textbf{Three analytical axes of EVE.}
\textbf{A:} local latent dimension \(k\), from the atomic case \(k=1\) to higher-dimensional latent spaces.
\textbf{B:} local capacity / operating regime, shaped by KL regularization and capacity control through a target band on \(\bar{\mu}^{\,2}\).
\textbf{C:} local autoregression in latent space, introducing temporal persistence at the neuron level.
Together, these axes define the design space studied.}
\label{fig:eve_axes}
\end{figure}

Figure~\ref{fig:eve_axes} summarizes the three analytical axes of EVE.

\paragraph{Contributions.}
\begin{enumerate}
    \item We define \emph{EVE} as a local variational computational primitive: a neuron with an explicit prior, an amortized posterior and unit-level variational regularization, whose atomic form is \(k=1\) and whose natural extension is to higher latent dimensionalities \(k=2,3,\dots\).
    \item We introduce a unit-centered framework for studying the neuron's internal dimensions, together with local diagnostics and control variables---including effective KL, latent energy, band occupancy, drift and collapse indicators---that make its operating regime observable and, to some extent, controllable.
    \item We provide an empirical study in selected forecasting and tabular settings showing that varying neuron-level latent dimensionality induces distinct learned regimes and we formalize a local autoregressive extension that introduces temporal persistence at the unit level as a natural third axis of analysis.
\end{enumerate}

\section{Related Work}
\label{sec:related}

\paragraph{Positioning logic.}
Our proposal connects several established lines of work while remaining distinct from each of them.
Throughout the paper, the primary meaning of \emph{dimensions} is the neuron's internal latent dimensionality \(k\).
In this section, however, we use \emph{dimension}, \emph{context} and \emph{time} as a positioning device:
\emph{dimension} concerns where uncertainty and inference are located,
\emph{context} concerns what is locally observable and controllable,
and \emph{time} concerns where probabilistic dynamics is introduced.

\paragraph{Dimension.}
Bayesian neural networks place uncertainty on the weights, while the computational unit itself remains essentially scalar \citep{blundell2015weight}. Standard uncertainty - quantification tools such as calibration and out - of - distribution detection also mainly act at the level of predictions and outputs \citep{guo2017calibration,hendrycks2017baseline}. VAEs, by contrast, place variational inference at the center of the model through a prior, an amortized posterior, an ELBO and reparameterization \citep{kingma2014vae,rezende2014stochastic}. However, this inference is typically carried by global or per-datum latents, while the decoder remains composed of deterministic units. Earlier stochastic-network frameworks, such as \emph{wake--sleep} and gradient estimators for stochastic units (REINFORCE, relaxations, VIMCO), introduce distributed stochastic computation \citep{hinton1995wakesleep,williams1992reinforce,bengio2013stochasticneurons,jang2017gumbel,mnih2016vimco}, but they do not formulate the unit itself as an explicit local variational inference process with its own prior, posterior and unit-level variational objective.

\paragraph{Context.}
Local objectives and layer-wise criteria have long been used to stabilize or decompose learning \citep{hinton2006dbn,bengio2007greedy,nokland2019local,jaderberg2017dni}, but they do not in themselves give the unit a probabilistic semantics. Likewise, calibration and OOD detection provide useful diagnostics, but primarily at the output level \citep{guo2017calibration,hendrycks2017baseline}. EVE differs in that the neuron comes with internal diagnostics and local constraints: the band on \(\mu^2\), effective KL, out-of-band fractions, drift and collapse indicators define a regime of local control over both capacity and internal behavior. In this sense, EVE does not only introduce a local latent; it makes that latent locally measurable and adjustable.

\paragraph{Time.}
Latent-variable models for sequences introduce stochastic dynamics
through time, typically via global latent states, as in VRNN-style
models or deep Kalman filter variants \citep{chung2015vrnn,krishnan2015dkf}. EVE differs in the placement of dynamics: we focus on per-unit dynamics through a local autoregressive prior. The point is not only to add probabilistic dynamics, but to place that dynamics at the micro level and tie it to local stability and capacity constraints.

\paragraph{Summary.}
BNNs place uncertainty in the weights \citep{blundell2015weight}; standard VAEs place it in global latents and optimize a global ELBO \citep{kingma2014vae,rezende2014stochastic}; stochastic units introduce local noise without turning the unit into an explicit variational inference process \citep{williams1992reinforce,bengio2013stochasticneurons,jang2017gumbel,mnih2016vimco}; sequential latent models place dynamics mainly at the macro level \citep{chung2015vrnn,krishnan2015dkf}. EVE, by contrast, combines a per-neuron latent, an explicit prior and posterior, unit-level variational regularization, local capacity diagnostics and constraints and autoregressive dynamics. It differs along the three axes that structure this paper: \textbf{dimension}, \textbf{context} and \textbf{time}.

\section{Method}
\label{sec:method}

\paragraph{Principle.}
We formalize the variational neuron as a locally parameterized probabilistic unit with explicit neuron-level latent structure.
The primary design axis studied is its internal latent dimensionality \(k\).
We then introduce two additional structural axes---\emph{context} and \emph{time}---to study how different values of \(k\) behave.
Here, \emph{context} refers to local capacity, that is, the amount of contextual information effectively carried by the latent state and \emph{time} refers to local persistence introduced through autoregressive dynamics.
These three axes are combined in a training objective whose neuron-level probabilistic structure is monitored and partially controlled through explicit diagnostics and constraints:

\subsection{Dimension: a local variational neuron}
\label{sec:method_dimension}

Given an input \(h\), an EVE neuron carries an internal latent variable \(z\in\mathbb{R}^k\), a standard Gaussian prior
\begin{equation}
p(z)=\mathcal{N}(0,I_k),
\end{equation}
and a diagonal amortized posterior
\begin{equation}
q_\phi(z\mid h)=\mathcal{N}\!\big(\mu_\phi(h),\mathrm{Diag}(\sigma_\phi(h)^2)\big).
\end{equation}
Sampling is performed via reparameterization,
\begin{equation}
z=\mu_\phi(h)+\sigma_\phi(h)\odot\varepsilon,
\qquad
\varepsilon\sim\mathcal{N}(0,I_k),
\end{equation}
after which the neuron emits a local representation or activation
\begin{equation}
a = g_\theta(z,h).
\end{equation}

At the conceptual level, this motivates the following generic unit-level variational formulation:
\begin{equation}
\label{eq:unit_elbo_compact}
\mathcal{L}_{\mathrm{unit}}(h)
=
\underbrace{\mathbb{E}_{q_\phi(z\mid h)}\!\left[-\log p_\theta(a^{*}\mid z,h)\right]}_{\text{local fit}}
+
\underbrace{\beta\,\mathrm{KL}\!\left(q_\phi(z\mid h)\,\|\,p(z)\right)}_{\text{variational regularization}}.
\end{equation}

Equation~\eqref{eq:unit_elbo_compact} gives a generic unit-level variational formulation. In the experimental instantiation studied, however, it is not optimized as a standalone local ELBO, since neurons do not receive a separate supervised target \(a^*\).

Instead, their stochastic local representations are composed through the layer readout and trained through the downstream task loss.
Accordingly, the local fit term is realized indirectly through the contribution of neuron-level latent representations to the global prediction \(\hat y\), while the variational part remains explicit through the KL term.
This yields the practical objective:
\begin{equation}
\label{eq:global_task_kl}
\mathcal{L}
=
\mathcal{L}_{\mathrm{task}}(\hat y,y)
+
\beta\,\mathrm{KL}_{\mathrm{mean}},
\end{equation}
with
\begin{equation}
\label{eq:kl_mean_method}
\mathrm{KL}_{\mathrm{mean}}
=
\frac{1}{N}\sum_{i=1}^{N}\mathrm{KL}\!\left(q_i(z_i\mid h)\,\|\,p(z_i)\right),
\end{equation}
where \(N\) denotes the number of neurons in the layer.
Averaging across neurons stabilizes the scale of the variational term as width increases.

The case \(k=1\) is the atomic form of the neuron, while \(k>1\) corresponds to higher latent dimensionalities.
The deterministic limit is recovered as \(\sigma_\phi(h)\to 0\).

Equations~\eqref{eq:global_task_kl}--\eqref{eq:kl_mean_method} define the experimental instantiation used throughout this paper.
Our empirical claims concern a locally parameterized probabilistic unit embedded in end-to-end optimization: the neuron carries explicit prior / posterior structure and neuron-level regularization, while supervision is provided through the downstream task objective.

\paragraph{Aggregation and readout.}
For a layer of \(N\) neurons, we aggregate latent statistics through a normalized readout:
\begin{equation}
\label{eq:readout_norm_method}
\hat y=\frac{w^\top m(h)}{\sqrt{N}}+b_0,
\end{equation}
where \(m(h)\) denotes the vector of local representations used in the readout (typically the means \(\mu_i(h)\), or optionally the samples \(z_i\)). The factor \(1/\sqrt{N}\) stabilizes the scale at large width.

\subsection{Context: local capacity and constraints}
\label{sec:method_context}

The second axis concerns the neuron's local capacity, defined as the amount of contextual information effectively carried by its latent state. For a neuron of dimension \(k\), we define the mean latent energy as
\begin{equation}
\label{eq:mu2_k}
\bar{\mu}^{\,2}
=
\mathbb{E}_h\!\left[\frac{1}{k}\|\mu_\phi(h)\|_2^2\right].
\end{equation}
This quantity acts as a proxy for capacity. Indeed, for a diagonal posterior,
\begin{equation}
\label{eq:kl_diag_k}
\mathrm{KL}\!\left(q_\phi(z\mid h)\,\|\,p(z)\right)
=
\frac{1}{2}\sum_{r=1}^{k}
\left(
\mu_r(h)^2+\sigma_r(h)^2-\log \sigma_r(h)^2-1
\right).
\end{equation}
When the variances remain close to a fixed or weakly varying regime, the KL is approximately affine in \(\|\mu_\phi(h)\|_2^2\), so that \(\bar{\mu}^{\,2}\) provides a directly interpretable proxy for the neuron's activity and information budget.

We then impose a capacity band
\begin{equation}
\ell \;\le\; \bar{\mu}^{\,2} \;\le\; u,
\end{equation}
which defines the admissible regime of the unit. In practice, we use either a soft penalty
\begin{equation}
\label{eq:band_loss}
\mathcal{L}_{\mathrm{band}}
=
\big(\ell-\bar{\mu}^{\,2}\big)_+^2
+
\big(\bar{\mu}^{\,2}-u\big)_+^2,
\end{equation}
or hard control by multiplicative projection after each update. If the parameters of the mean head are denoted \((A_\mu,b_\mu)\), we apply
\begin{equation}
\label{eq:projection_k}
(A_\mu,b_\mu)\leftarrow s(A_\mu,b_\mu),
\end{equation}
with
\[
s=
\begin{cases}
\sqrt{\ell/\bar{\mu}^{\,2}} & \text{if } \bar{\mu}^{\,2}<\ell,\\[2mm]
\sqrt{u/\bar{\mu}^{\,2}} & \text{if } \bar{\mu}^{\,2}>u,\\[2mm]
1 & \text{otherwise.}
\end{cases}
\]
This projection brings the latent energy exactly back into the admissible band whenever it leaves it.

To that end, we track several internal observables: \(\mathrm{KL}_{\mathrm{mean}}\), an effective KL \(\mathrm{KL}_{\mathrm{eff}}=\max(\mathrm{KL}-\tau,0)\), the energy \(\bar{\mu}^{\,2}\), band-violation rates (\emph{low/high rates}), the amplitude of projection corrections, as well as a reparameterization proxy \(\mathbb{E}\|z-\mu\|_1/k\).

\subsection{Time: local autoregressive dynamics}
\label{sec:method_time}

The third axis concerns persistence through time. For a sequence \((h_t)_{t=1}^T\), we introduce an autoregressive dynamics at the level of the neuron itself. In its explicit form, the prior becomes
\begin{equation}
\label{eq:ar_prior_k}
p(z_t\mid z_{t-1})
=
\mathcal{N}(\phi z_{t-1},\sigma_{\mathrm{AR}}^2 I_k),
\qquad
\phi = \exp(-\Delta t/\tau),
\end{equation}
where \(\tau\) plays the role of a local time scale. In practice, we use a causal form that can be implemented as a penalty on the posterior means:
\begin{equation}
\label{eq:ar_penalty_k}
\mathcal{L}_{\mathrm{AR}}
=
\frac{1}{T-1}\sum_{t=2}^{T}
\left\|
\mu_t-\phi\,\mu_{t-1}
\right\|_2^2.
\end{equation}

This extension does not change the meaning of \(k\): it adds a second control axis, orthogonal to capacity. The band on \(\bar{\mu}^{\,2}\) regulates energetic amplitude and the amount of contextual information effectively carried by the neuron; the AR dynamics regulates persistence and the speed at which this internal state evolves. In our terminology, \emph{context} and \emph{time} are thus two structural properties added to the neuron's latent dimensionality.

\subsection{Unified view and optimization}
\label{sec:method_unified}

These three axes are combined in a single training objective, whose local probabilistic structure is made observable and controllable through explicit diagnostics and constraints:
\begin{equation}
\label{eq:method_full_loss}
\mathcal{L}
=
\mathcal{L}_{\mathrm{task}}(\hat y,y)
+
\beta\,\mathrm{KL}_{\mathrm{mean}}
+
\lambda_{\mathrm{band}}\,\mathcal{L}_{\mathrm{band}}
+
\alpha_{\mathrm{AR}}\,\mathcal{L}_{\mathrm{AR}},
\end{equation}
optionally complemented by hard projection onto the admissible band when the controller is activated.

This expression makes the logic of the paper explicit:
\begin{itemize}
    \item \textbf{dimension}: the choice of \(k\), together with \(\mathrm{KL}_{\mathrm{mean}}\), defines the neuron's variational regime;
    \item \textbf{context}: \(\mathcal{L}_{\mathrm{band}}\) constrains local capacity through latent energy;
    \item \textbf{time}: \(\mathcal{L}_{\mathrm{AR}}\) constrains local persistence.
\end{itemize}

Optimization uses Adam or AdamW.
In practice, we use one Monte Carlo sample per forward pass.
We apply standard safeguards:
\texttt{isfinite} checks, gradient clipping,
norm monitoring, and multi-seed evaluation.
Operating points are selected on validation data by combining
predictive quality with internal regime diagnostics.

\section{Experiments}
\label{sec:experiments}

\paragraph{Goal, scope, setup and reporting.}
We study the variational neuron as a \emph{probabilistic computational} primitive with explicit internal dimensions across long-horizon forecasting, tabular regression and a controlled low-dimensional benchmark on \textsc{MNIST}. For forecasting, we use the standard LongHorizon protocol with input length \(336\) and horizon \(H=96\), compare \texttt{homeo}, \texttt{projOFF} and \texttt{projON}, report external performance by test MSE and track KL, latent energy \(\mu^2\) and the out-of-band fraction \(\textit{out}=\mathrm{frac}_{low}+\mathrm{frac}_{high}\); unless otherwise stated, values are mean \(\pm\) standard deviation over seeds. Main evidence consists of: (i) a global forecasting result showing that an internal regime variable is statistically informative, (ii) a dedicated compact-dimension study on \textsc{ECL} and (iii) two favorable low-dimensional tabular cases, \textsc{ConcreteStrength} and \textsc{CaliforniaHousing}; the remaining datasets extend the empirical picture by testing scope, consistency and boundary conditions.

\subsection{Main evidence I: internal regime variables are statistically informative}
\label{sec:out_proxy}

Across \(N=60\) dataset-normalized forecasting runs, the fraction of units outside the operating band, \(\textit{out}\), correlates positively with test MSE (\(r=0.567\), \(p=2.3 \times 10^{-6}\)). As band violations increase, predictive quality degrades on average, making \(\textit{out}\) a statistically informative proxy for regime quality; dataset-level forecasting cases in Section~\ref{sec:auxiliary_support} further clarify where this signal is most useful.

\subsection{Main evidence II: ECL compact-dimension search}
\label{sec:ecl_compact_search}

We next test whether increasing latent dimensionality on \textsc{ECL} yields a better \emph{probabilistic operating regime} under a controlled shared budget.

\begin{figure}[t]
    \centering
    \includegraphics[width=\linewidth]{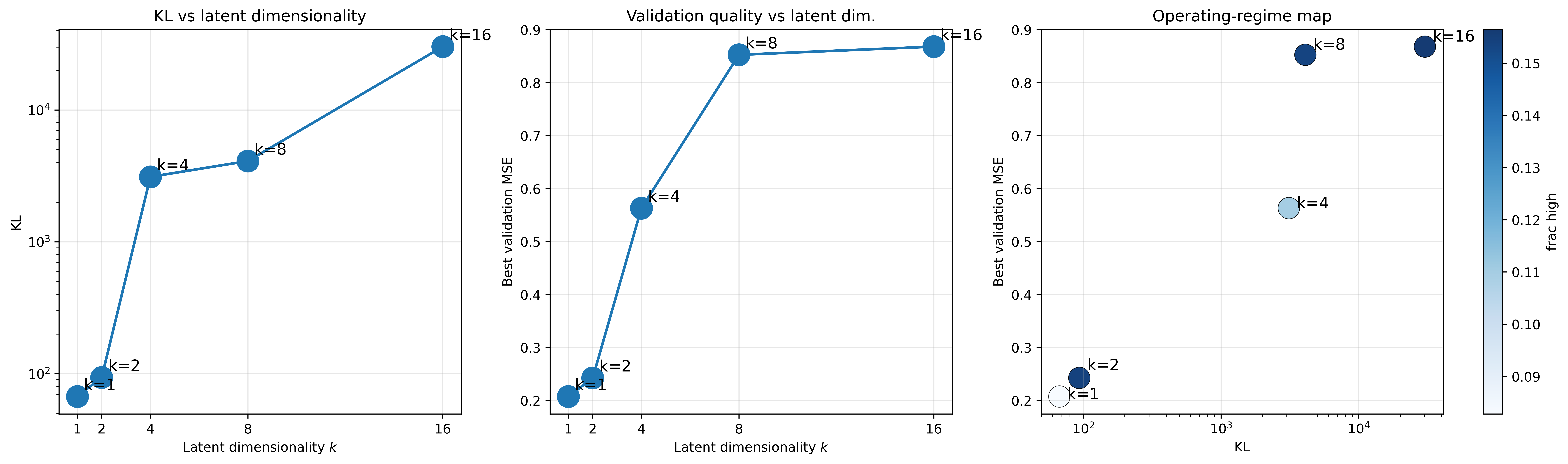}
\caption{\textbf{ECL compact-dimension search.}
Left: KL versus latent dimensionality \(k\).
Middle: best validation MSE versus \(k\).
Right: operating-regime map with KL on the horizontal axis, best validation MSE on the vertical axis and color indicating \textit{frac high}.
The compact case \(k=1\) yields the most favorable regime under the shared-budget comparison.}
    \label{fig:ecl_compact_regime}
\end{figure}

\begin{table}[t]
\centering
\small
\setlength{\tabcolsep}{5pt}
\renewcommand{\arraystretch}{1.06}
\caption{\textbf{ECL compact-dimension search.}
Best observed configuration for each latent dimension during the first shared-budget comparison stage.}
\label{tab:ecl_compact_search}
\begin{tabular}{c c c c c c}
\toprule
\boldmath$k$ & \textbf{best val MSE\(\downarrow\)} & \textbf{best score\(\downarrow\)} & \textbf{inside mass\(\uparrow\)} & \textbf{frac high\(\downarrow\)} & \textbf{KL\(\downarrow\)} \\
\midrule
1  & \(0.2073\) & \(0.2179\) & \(0.6752\) & \(0.0828\) & \(67.1\) \\
2  & \(0.2424\) & \(0.2778\) & \(0.6396\) & \(0.1541\) & \(93.6\) \\
4  & \(0.5632\) & \(0.5643\) & \(0.6938\) & \(0.1099\) & \(3111.3\) \\
8  & \(0.8527\) & \(2.0438\) & \(0.3380\) & \(0.1535\) & \(4098.4\) \\
16 & \(0.8682\) & \(1.6241\) & \(0.3943\) & \(0.1565\) & \(30227.3\) \\
\bottomrule
\end{tabular}
\end{table}

Figure~\ref{fig:ecl_compact_regime} and Table~\ref{tab:ecl_compact_search} show that increasing \(k\) sharply raises KL and degrades validation quality: from \(k=1\) to \(k=16\), KL rises from \(67.1\) to \(30227.3\), while best validation MSE degrades from \(0.2073\) to \(0.8682\). Under this shared-budget comparison, the most favorable regime is compact, \(k=1\). The selected compact regime was then carried forward in the subsequent pipeline, yielding \(\text{best val MSE}=0.2043\) and \(\text{best score}=0.2120\). A matched AR ON/OFF ablation further shows that temporal structure is itself a genuine regime variable whose effect depends on the learned operating regime: enabling AR degrades test MSE from \(0.217 \pm 0.003\) to \(0.461 \pm 0.082\), raises \(\texttt{ar\_share}\) from \(0.000\) to \(0.367 \pm 0.219\) and raises \(\textit{out}\) from \(0.393 \pm 0.007\) to \(0.601 \pm 0.065\).

\subsection{Main evidence III: favorable low-dimensional tabular regimes}
\label{sec:main_tabular_cases}

\subsubsection{ConcreteStrength}
\label{sec:concrete_main}

\textsc{ConcreteStrength} is a favorable tabular case in which a low-dimensional variational configuration clearly outperforms DET.

\begin{table}[t]
\centering
\footnotesize
\setlength{\tabcolsep}{4pt}
\renewcommand{\arraystretch}{1}
\caption{\textbf{ConcreteStrength main results.}
Comparison between the deterministic baseline and one reported low-dimensional variational configuration.}
\label{tab:concrete_main}
\begin{tabular}{p{1.7cm} c c c c}
\toprule
\textbf{Model} & \textbf{Test MSE\(\downarrow\)} & \textbf{Test MAE\(\downarrow\)} & \textbf{Test CRPS\(\downarrow\)} & \textbf{Test NLL\(\downarrow\)} \\
\midrule
DET & \(0.3049 \pm 0.0459\) & \(0.4309 \pm 0.0342\) & \(0.3202 \pm 0.0285\) & \(1.1602 \pm 0.1987\) \\
\(k=1\), \texttt{homeo} & \(\mathbf{0.1977 \pm 0.0190}\) & \(\mathbf{0.3371 \pm 0.0134}\) & \(\mathbf{0.2460 \pm 0.0107}\) & \(\mathbf{0.6729 \pm 0.0766}\) \\
\bottomrule
\end{tabular}
\end{table}

The reported \(k=1\), \texttt{homeo} regime outperforms DET on all reported metrics. Across the broader sweep, the global optimum is \(k=2\), \texttt{projOFF}, with test MSE \(0.1934 \pm 0.0240\), while the selected \(k=1\), \texttt{homeo} model remains very close and provides a more centered internal regime.

\subsubsection{CaliforniaHousing}
\label{sec:california_main}

\textsc{CaliforniaHousing} provides a second favorable tabular case.

\begin{table}[t]
\centering
\small
\setlength{\tabcolsep}{5pt}
\renewcommand{\arraystretch}{1.06}
\caption{\textbf{CaliforniaHousing: deterministic baseline versus a low-dimensional variational regime.}}
\label{tab:california_main}
\begin{tabular}{l c c c c}
\toprule
\textbf{Model} & \textbf{Test MSE\(\downarrow\)} & \textbf{KL} & \textbf{inside mass\(\uparrow\)} & \textbf{frac too low\(\downarrow\)} \\
\midrule
DET & \(0.2812 \pm 0.0485\) & -- & -- & -- \\
\(k=1\), \texttt{projON} & \(\mathbf{0.2598 \pm 0.0254}\) & \(46.64 \pm 2.21\) & \(0.906 \pm 0.044\) & \(0.076 \pm 0.043\) \\
\bottomrule
\end{tabular}
\end{table}

The \(k=1\), \texttt{projON} regime improves over DET while remaining structurally well behaved. In the broader sweep, \(k=2\), \texttt{projON} gives the best validation score and \(k=4\), \texttt{projON} the best observed test MSE (\(0.2510 \pm 0.0150\)), with more strained latent diagnostics, including larger KL and lower inside mass.

\subsection{Supporting evidence: scope, limits,
 additional cases}
\label{sec:auxiliary_support}

\paragraph{Forecasting support.}
The auxiliary forecasting datasets support the same structural reading while clarifying its scope. On \textsc{Weather}, \texttt{homeo} is the most favorable reported EVE regime, with \(0.183 \pm 0.009\) test MSE and \(\textit{out}=0.573 \pm 0.022\), versus \(0.193 \pm 0.005\) and \(0.772 \pm 0.012\) for \texttt{projOFF} and \(0.191 \pm 0.008\) and \(0.783 \pm 0.012\) for \texttt{projON}; it is also best in MSE for \(5/5\) seeds, while DET is slightly stronger externally in the dedicated DET-vs-EVE comparison (\(0.177 \pm 0.012\) versus \(0.189 \pm 0.011\)). On \textsc{Exchange}, the selected \texttt{homeo} regime reaches \(0.180 \pm 0.037\) test MSE with \(\mu^2 = 0.150 \pm 0.011\) and \(\textit{out}=0.403 \pm 0.036\); relative to \texttt{projOFF}, it reduces inter-seed MSE variance by about \(2.9\times\), yields lower \(\textit{out}\) in \(5/5\) seeds and improves regime stability, while DET remains stronger overall (\(0.099 \pm 0.008\) test MSE). On \textsc{TrafficL}, \texttt{projOFF} and \texttt{projON} remain close at \(0.659 \pm 0.006\) and \(0.669 \pm 0.008\) test MSE, with \(\textit{out}=0.379 \pm 0.009\) and \(0.379 \pm 0.014\), whereas \texttt{homeo} yields \(\textit{out}=0.530 \pm 0.014\) and correspondingly higher test MSE, \(1.117 \pm 0.005\); DET is again stronger (\(0.371 \pm 0.001\)). These cases show that \(\textit{out}\) is informative and that the effect of control is dataset-dependent.

\paragraph{Additional temporal evidence.}
A matched AR ON/OFF comparison confirms that local autoregressive dynamics is a genuine neuron-level regime variable whose effect is dataset-dependent: on \textsc{Exchange}, AR ON slightly improves test MSE over AR OFF (\(0.180 \pm 0.037\) vs \(0.190 \pm 0.049\)) with \(\texttt{ar\_share}=0.047 \pm 0.005\), whereas on \textsc{ECL} the same switch degrades performance, as reported above.

\paragraph{Additional structural tabular cases.}
\textsc{ProteinStructure} shows strong controller dependence: under \texttt{homeo}, the best final regime is obtained at \(k=1\), with best validation neg-ELBO \(1.122 \pm 0.011\), best validation NLL \(1.092 \pm 0.011\) and best epoch \(18.1 \pm 2.2\); moving to \(k=2\) degrades the fit to \(1.265 \pm 0.020\) in best validation neg-ELBO and \(1.134 \pm 0.019\) in best validation NLL, with best epoch \(6.9 \pm 1.0\). More broadly, the tuning scores under \texttt{homeo} deteriorate from \(1.131\) at \(k=1\) to \(1.282\), \(1.621\), \(2.553\) and \(4.095\) at \(k=2,4,8,16\), whereas under \texttt{projOFF} they remain \(1.129\), \(1.142\) and \(1.171\) at \(k=1,2,4\). A dedicated rerun of \(k=4\), \texttt{projOFF} reaches \(1.1811 \pm 0.0106\) best validation neg-ELBO, \(1.1021 \pm 0.0127\) best validation NLL, \(1.1759 \pm 0.0082\) test neg-ELBO, \(1.0971 \pm 0.0077\) test NLL and \(0.5253 \pm 0.0080\) test MSE. \textsc{BostonHousing} provides an informative contrast case: the best reported variational regime is \texttt{projON} at \(k=2\), with test MSE \(0.381\), test NLL \(1.578\) and test CRPS \(0.307\); under \texttt{projON}, test MSE moves from \(0.403\) at \(k=1\) to \(0.381\) at \(k=2\), remains \(0.394\) at \(k=4\) and deteriorates to \(0.473\) at \(k=8\) and \(0.620\) at \(k=16\), whereas DET remains better overall (\(0.305\) test MSE, \(1.203\) test NLL, \(0.284\) test CRPS). \textsc{EnergyEfficiency} adds another favorable low-dimensional case: the best grid point is \(k=4\), \texttt{projOFF}, with test MSE \(0.209\) and test NLL \(0.732\), while the conference selection \(k=2\), \texttt{projON} reaches best validation MSE \(0.189\); in the final aligned comparison, \(\texttt{nvae\_best}\) reaches test MSE \(0.213\), test NLL \(0.755\) and CRPS \(0.217\), versus DET at \(0.258\), \(0.974\) and \(0.214\).

\paragraph{Controlled low-dimensional probabilistic benchmark.}
\textsc{MNIST} results support the same structural reading at very small latent dimension. Under validation-MSE selection and \(6\) seeds, enabling structural projection improves both \(k=1\) and \(k=2\) regimes: for \(k=1\), validation MSE decreases from \(0.02200 \pm 0.00043\) under \texttt{projOFF} to \(0.01884 \pm 0.00038\) under \texttt{projON}, test MSE from \(0.02163 \pm 0.00046\) to \(0.01857 \pm 0.00041\) and test pinball from \(0.03979 \pm 0.00054\) to \(0.03824 \pm 0.00049\); for \(k=2\), validation MSE decreases from \(0.02256 \pm 0.00071\) to \(0.01881 \pm 0.00036\), test MSE from \(0.02219 \pm 0.00069\) to \(0.01854 \pm 0.00039\) and test pinball from \(0.03941 \pm 0.00062\) to \(0.03806 \pm 0.00047\). The mean predictive standard deviation remains stable across settings (approximately \(0.23\)) and the difference between \(k=1\) and \(k=2\) under \texttt{projON} is negligible (\(0.01857\) versus \(0.01854\) in test MSE).

\subsection{Synthesis}
\label{sec:exp_synthesis}

The experiments support three conclusions: \(\textit{out}\) is statistically linked to predictive degradation across runs; latent dimensionality acts as a structural variable of the neuron, with the controlled \textsc{ECL} study identifying \(k=1\) as the most favorable operating point; and low-dimensional variational regimes can be empirically competitive, as shown by \textsc{ConcreteStrength} and \textsc{CaliforniaHousing}. Auxiliary datasets preserve the same picture while clarifying its dataset dependence: \textsc{Weather}, \textsc{Exchange} and \textsc{TrafficL} show that control can strongly reshape long-horizon regime quality, with gains that depend on the dataset relative to DET; \textsc{ProteinStructure} and \textsc{BostonHousing} show that the effect of latent dimensionality is non-monotonic and controller-dependent; \textsc{EnergyEfficiency} adds another favorable low-dimensional tabular case; and \textsc{MNIST} shows that structural projection can substantially improve predictive and probabilistic behavior even when moving from \(k=1\) to \(k=2\) yields almost no additional benefit. The main message is that varying the neuron's latent dimensionality \(k\) induces distinct regimes characterized by control, internal activity, band occupancy, and, preliminarily, local temporal dynamics.

\section{Discussion}
\label{sec:discussion}
This paper studies the variational neuron as a locally parameterized probabilistic object with explicit internal dimensions. Empirically, neuron-level dimensions induce distinct, measurable regimes, and some internal variables are informative about downstream quality. Taken together, these findings position EVE as a framework for making neuron-level probabilistic regimes experimentally accessible and partially controllable across settings.

A natural next step is to allow heterogeneous neuron-level dimensionalities within the same network, making it possible to study structured hierarchies of local uncertainty and capacity across units.

More broadly, the results support studying neuron-level probabilistic structure as an experimental object in its own right across architectures and tasks. Extending the benchmarks, strengthening the analysis of latent dynamics and band occupancy and evaluating predictive uncertainty more systematically are direct directions for future work.

\section{Conclusion}
\label{sec:conclusion}
This paper introduced EVE as a local variational computational primitive and studied the dimensions of a variational neuron primarily through its internal latent dimensionality \(k\). Across the settings considered here, the results show that neuron-level regime variables are measurable, that latent dimensionality acts as a structural variable rather than a cosmetic capacity parameter and that favorable low-dimensional EVE regimes can be both interpretable and useful in practice. In this sense, exploring the dimensions of a variational neuron makes local probabilistic computation experimentally accessible and turns neuron-level uncertainty into a measurable design space rather than a hidden implementation detail.

\bibliographystyle{plainnat}
\bibliography{references}

\end{document}